\documentclass[letterpaper, 10 pt, conference]{oldIEEEFormatting/ieeeconf}  %

\IEEEoverridecommandlockouts                              %

\usepackage{graphicx} %
\usepackage{amsmath} %
\usepackage{amssymb}  %

\usepackage{setspace}

\usepackage{verbatim}
\usepackage{afterpage}
\usepackage[colorlinks]{hyperref}
\usepackage[usenames, dvipsnames]{color}

\usepackage[normalem]{ulem}

\newcommand{\refFig}[1]{(Figure \ref{#1})}

\newcommand{\figRef}[1]{(Figure \ref{#1})}
\newcommand{\figRefAnnotate}[2]{(Figure \ref{#1}#2)}

\newcommand{\IOC}{IOC\ }

\newcommand{\edit}[1]{#1}

\newcommand{\xVect}{\ensuremath{\mathbf{x}}}
\newcommand{\weights}{\ensuremath{\mathbf{w}}}

\newcommand{\trajOpt}{\ensuremath{\mathbf{x}^*}}

\newcommand{\weightCandidate}{\ensuremath{w}}

\title{\LARGE \bf
\edit{A Robustness Analysis of Inverse Optimal Control of Bipedal Walking}
}

\author{John R. Rebula$^{1}$, Stefan Schaal$^{2}$, James Finley$^{3}$, Ludovic Righetti$^{1, 4}$%
\thanks{
This work was supported by
 New York University, the National Science Foundation (grant CMMI-1825993), NIH NICHD Award Number R01HD091184, National Science Foundation grants IIS-1205249, IIS-1017134, EECS-0926052, the Office of Naval Research, the Okawa Foundation, the Max-Planck-Society and the European Research Council (ERC) under the European Union Horizon 2020 research and innovation programme (grant agreement No 637935 and European Research Councils grant No 637935).
}
\thanks{$^{1}$John R. Rebula and Ludovic Righetti are with The Max Planck Institute for Intelligent Systems of Tuebingen, Germany
    {\tt\footnotesize jrebula@umich.edu}}%
\thanks{$^{2}$Stefan Schaal is with Google X of Mountain View, CA}
\thanks{$^{3}$James Finley is with The University of Southern California of Los Angeles, USA}
\thanks{$^{4}$Ludovic Righetti is with New York University of New York, USA}}

\begin{document}

\maketitle

\begin{abstract}

Cost functions have the potential to provide compact and understandable generalizations of motion. The goal of Inverse Optimal Control (IOC) is to analyze an observed behavior which is assumed to be optimal with respect to an unknown cost function, and infer this cost function. Here we develop a method for characterizing cost functions of legged locomotion, with the goal of representing complex humanoid behavior with simple models. To test this methodology we simulate walking gaits of a simple 5 link planar walking model which optimize known cost functions, and assess the ability of our IOC method to recover them. In particular, the IOC method uses an iterative trajectory optimization process to infer cost function weightings consistent with those used to generate a single demonstrated optimal trial. We also explore sensitivity of the IOC to sensor noise in the observed trajectory, imperfect knowledge of the model or task, as well as uncertainty in the components of the cost function used. With appropriate modeling, these methods may help infer cost functions from human data, yielding a compact and generalizable representation of human-like motion for use in humanoid robot controllers, as well as providing a new tool for experimentally exploring human preferences.

\end{abstract}

\section{Introduction}

One approach to generating highly capable robotic behavior is to take inspiration from natural examples of animals with similar morphologies. For example, humans provide an excellent model for bipedal locomotion as they are highly maneuverable and can reliably traverse complex terrain. While human walking is well studied, a question remains of how to transfer the body of knowledge of human gait into robust, general controllers for humanoid robots.
While biomechanists typically measure and describe existing gait, the roboticist generates gait from scratch.
In this sense, the biomechanist and roboticist approach gait from opposite directions.

One important area of overlap between biomechanics and robotics, however, is the use of simple mechanical walking models. Simple models such as the linear inverted pendulum model (LIPM) have been used both to describe human stability \cite{hof_extrapolated_2008}, as well as generate robust stepping behavior for robots \cite{koolen_capturability-based_2012part1}. One advantage of such models is their relatively low dimensionality, making them amenable to optimization based analysis. For example, the LIPM can be used to predict an appropriate place to step for the simplified model, and the more complex robot can step there \cite{pratt_capturability-based_2012}, or this can be used as a starting location for additional refinement to maximize stability \cite{rebula_learning_2007}. More complicated models, such as the compass walker \cite{usherwood_compass_2008} and ballistic walker \cite{mochon_ballistic_1980} add hip and knee joints, and allow reasoning about joint torques in simplified gait, while still remaining tractable for trajectory optimization methods. However, even when a simple model is chosen as a basis for gait generation, the question of an appropriate cost function remains. %
While cost functions for robot controllers are typically hand-tuned, we would like to find simple model gaits which optimize human cost functions, potentially inheriting some desirable aspects of human gait.

A central question is how to identify a cost function given behavior observations. One approach is for investigators to try different cost functions, choosing one which results in a robot gait most closely resembling human gaits.
However, this may require the designer to laboriously tune a number of gains with little a priori knowledge of the potentially vast search space.
Our goal is to determine a cost function computationally from observed behavior. This process is called inverse optimal control (IOC), as instead of using a model cost function
to calculate optimal behavior, as in optimal control, we must use a model and \edit{measurements of} a behavior to determine the cost function. This cost function
allows generalization from the demonstrated behavior using the predictive model, while also inheriting its simplifying assumptions.

In a realistic application of IOC, such as when characterizing the cost function underlying a human behavior, the measurement of the optimal behavior will be subject to imperfect measurements, and the model used within the IOC procedure will necessarily be a simplification of the human. In addition, the cost function is composed of hypothetical cost components which must be chosen with care based on the experiment being performed. We seek to analyze the implications of these imperfections on the results of an IOC analysis.

\subsection{Previous work}

Under certain conditions, humans may be thought of as performing optimal control.
For example, a person in typical laboratory conditions, walking on a treadmill at a set speed for long periods of time, may choose a
gait which uses minimal effort to stay on the treadmill. Studies of human gait under such conditions have found that subjects choose many measurable
aspects of their gait to minimize metabolic energy expenditure. For example, healthy subjects walking under conditions which alter their step width \cite{donelan_mechanical_2001}, step length \cite{zarrugh_optimization_1974}, or arm swinging \cite{collins_dynamic_2009}, require more energy than when using their preferred gait.
Even devices designed to add energy into gait may alter the gait away from the naturally chosen motion, and without careful timing generally lead to a more costly gait \cite{malcolm_simple_2013}.
While some metrics clearly factor into specific tasks, such as energy considerations during treadmill
walking, these motions are likely defined by many more subtle factors as well. For example, in jump landing trials, both mechanical energy (one contributor
to metabolic energy cost) and collision impulse (which might be dangerous or painful) are both considered \cite{zelik_mechanical_2012}. %
Based on such experimental observations of humans acting as optimal controllers, we will seek to develop a method to determine the underlying cost functions from observed human behavior.

Optimization is one method used in humanoid control to select one option among many possibilities which satisfy a task. Some methods calculate a long term trajectory \cite{posa_direct_2013} to perform overall tasks such as traversing terrain. Optimization is also used at a faster rate to resolve the redundancy associated with the many degrees of freedom of the robot to perform a somewhat more specified task, such as center of mass tracking at high feedback rates \cite{herzog_balancing_2014}. A typical cost minimized in these approaches involves multiple terms such as the actuation effort and the difference from a nominal pose. The specific weights chosen to combine these various components into an overall cost function are typically tuned by hand for a desirable result. A humanoid robot performing a task faces some of the same control challenges as humans performing that task. While the particular sensing, actuation, and computational hardware differ, they share some aspects, such as physics and task requirements, as well as fragility to impacts and energy limitations. Therefore, we consider humans behavior a good basis for humanoid robot cost functions.

\IOC has been used to resolve the issues of determining optimal policies and cost functions from demonstrations.
Cost functions have been inferred for humans performing reaching and manipulation tasks \cite{doerr_direct_2015}.
\IOC algorithms are also applied to grid-world type scenarios where the state space is partitioned and a policy is learned to navigate it based on demonstrations. Some of these methods have difficulty extending to higher dimensional spaces \cite{doerr_direct_2015}, and some require multiple example trajectories. Algorithms have been developed which relax the need for a globally optimal example in favor of finding rewards locally along demonstrated paths \cite{levine2012Continuous}.
This has the advantage of not requiring a globally optimal demonstration. However, these works are largely focused on generating policies to recreate behavior, rather than the underlying cost function \cite{dvijotham_inverse_2010}. This is useful to predict human motions to facilitate human-robot interaction \cite{mainprice_goal_2016}.
It is also used in imitation learning where, for example, a human moves the robot through desired motions and \IOC infers a controller. Machine learning has been used to learn cost features automatically \cite{finn_guided_2016}. While useful for imitation learning, we are interested in learning general biomechanically relevant cost functions. Work on extracting biomechanically relevant costs from human upper body motions includes pointing game experiments with different reward and noise property conditions \cite{berniker_examination_2013}. Kinematic and force cost components considered did not generalize across conditions, whether because the cost components failed to capture the true cost or the subjects did not perform reliably.

Our implementation of \IOC is designed to handle the underactuation inherent in locomotion.
Underactuation represents fundamental constraints, so we choose an explicitly constrained optimization approach.
As we are interested in cyclic gait, we must also constrain the initial and final configuration to match.
Approaches to manipulator path planning often rely on unconstrained optimization approaches \cite{doerr_direct_2015}. It is unclear how to simply introduce nonlinear, impulsive constraints into many of the inverse reinforcement learning methods designed to learn policies based on state-space grids or Gaussian processes in high dimensions. \IOC has been used to estimate human cost functions in locomotion navigation in cluttered spaces \cite{Mombaur_From_2010}. \IOC based on constrained optimization has also been applied to joint level analysis of human walking
\cite{clever_inverse_2016} \edit{and transferring human motion to robot control through a simple model \cite{clever_humanoid_2018}}. In graphics applications, approximations softening contact constraints have been used to facilitate \IOC based on motion capture data to improve visual realism over conventional graphics methods \cite{liu_Learning_2005}.

\subsection{Our approach}

One open question in IOC in locomotion is how well the process can be trusted to produce usable results. For example, joint level IOC of human walking \cite{clever_inverse_2016}, is able to reproduce observed behaviors qualitatively well, but it is unclear how much trust can be placed in the weights found. It is important to know how well an IOC method disambiguates between different potential cost functions, and how much it is affected by imperfect modelling. One useful feature of an IOC process is the ability to produce the correct cost function under ideal conditions, for example using simulated data with perfectly known cost functions. Such analyses have been performed for methods applied \edit{to} manipulation tasks (e.g. \cite{finn_guided_2016}), but to date we have not found such an analysis for locomotion tasks. The closest work\edit{s} to ours \edit{are} an IOC analysis of experimental \edit{steady walking} data from several subjects \cite{clever_inverse_2016}
\edit{, and IOC of a simple model to transfer human walking to robots \cite{clever_humanoid_2018}. The steady walking human study results show} interesting clustering of cost functions from the subject data.
\edit{The simple model study shows how complex motions such as humans walking on stepping stones can be captured with a simple optimization model and transferred to robot behavior.} An ideal result from an IOC analysis is an understandable and generalizable cost function with measures of confidence. Such a cost function would then be useful for analyzing human behavior and transferring it to analogous robotic behavior.

As previously mentioned, one significant factor missing from the existing work on legged locomotion \IOC is an analysis of how robust the algorithms are. One possible reason for this is the difficulty in implementing a reliable high-dimensional constrained forward trajectory optimization.
To that end, we develop and test an \IOC algorithm for the purposes of robustly inferring cost functions. We will begin by testing recovery of a known cost function from a known model of walking, then we will consider the effects of imperfect measurements, an imperfect form of cost function, incorrect modelling, and incorrect task specifications (Figure \ref{fig:methodology}). This will illustrate the performance of the algorithm under more realistic experimental conditions.

\section{methods}

Here we detail our process of using \IOC to discover cost functions from observed optimal behavior.
We first develop a two dimensional linkage trajectory optimization methodology, with a focus on sparse constraints.
We then build a simple ballistic walking model with a torso using this methodology, and propose a set of cost components that we wish to minimize when choosing a gait. This allows generation of optimal gaits with a variety of different relative cost weightings.

We introduce our \IOC algorithm, along with a methodology to test its convergence under ideal conditions.
Since this work must be applicable to experimentally collected human data, the \IOC is tested under more realistic imperfect conditions. In particular, we test the impact of noisy measurements, incorrect model mass distribution, and an incorrect speed constraint.
Finally, we test the ability of the \IOC algorithm to handle the addition of incorrect components to the cost function, as a human subject's real cost components are unknown.

\newcommand{\tInT}{\ensuremath{t \in [1 \ldots T]}}
\newcommand{\tInTMinusOne}{\ensuremath{t \in [1 \ldots T-1]}}
\newcommand{\sInS}{\ensuremath{s \in [1 \ldots S]}}
\newcommand{\pInP}{\ensuremath{p \in [1 \ldots P]}}

\newcommand{\allStates}{\ensuremath{\vec{x}}}
\newcommand{\state}{\ensuremath{\vec{x}}}

\subsection{Segment based trajectory optimization}

Here we provide a brief overview of the optimization methodology used for the models presented in this paper. The trajectory optimization has the form
\begin{align*}
\min_{\allStates} w \cdot \phi(\allStates)  \hspace{1em}
\textrm{such that } \hspace{1em}
c(\allStates) \leq 0, \hspace{1em} ce(\allStates) = 0.
\end{align*}

The cost is a weighted sum of basis cost functions $\phi_i$, weighed by preference weights, $w_i$. The state of the system at time $t$ in the optimization, $x_t$, is considered throughout the trajectory, (\tInT). There are equality (ce) and inequality (c) constraints, some of which constrain the state to be physically representative of the model (e.g. the dynamics equations), and others enforce task constraints (e.g. repetitive gait).
\edit{The state trajectory, $x_t, \forall \tInT$, represents the behavior of the system. In principle the initial system state along with the actuation trajectories are sufficient to describe the behavior, but additional state variables allow simpler dynamics constraints.}

Here the model is represented as a set of $S$ two dimensional segments constrained with pin joints, with segment state $x_{s} = [x, z, \theta, \dot{x}, \dot{z}, \dot{\theta}]_{s,t}, \forall \sInS, \tInT$. Each segment $s$ has constraints:
\begin{align*}
F - m \cdot \left[\ddot{x}, \ddot{z}, \ddot{\theta}\right]_{s,t} = 0, \hspace{1em} \forall &\tInT \\
\state_{s, t+2} = \textrm{HermiteSimpsonIntegration} \Big( &
\state_{s, t}, \left[\ddot{x}, \ddot{z}, \ddot{\theta}\right]_{s,t}, \\
\state_{s, t+1}, \left[\ddot{x}, \ddot{z}, \ddot{\theta}\right]_{s,t+1}  \Big), & \forall t \in [1:2:(T-2)].
\end{align*}

\edit{
Hermite-Simpson integration assumes accelerations are quadratic and states are a 3rd order polynomial \cite{kelly_introduction_2017}.}

\subsubsection{Pin Joints}

Segments can be connected via pin joints. A pin joint adds a set of constraints throughout the trajectory, as well as forces to the segments it connects. The pin forces are found in the optimization, and so each pin joint $p$ has state
$
x_p = [f_x, f_z].
$
Adding a pin between segment $s_a$ and $s_b$, with position represented as $\vec{r}_{p,a}$ based on segment $a$, and $\vec{r}_{p,b}$ based on segment $b$ adds constraints:
\begin{equation*}
\vec{r}_{p,a,t} = \vec{r}_{p,b,t} \hspace{1em} \forall \tInT
\end{equation*}
as well as a final velocity matching constraint,
$
\dot{\vec{r}}_{p,a,T} = \dot{\vec{r}}_{p,b,T}.
$
The full state of the system is
\begin{align*}
&\allStates = [\Delta T, \state_{s,t} \state_{p,t}, [\ddot{x}, \ddot{z}, \ddot{\theta}]_{s,t}]  \\
&\forall \sInS, \forall \tInT, \forall \pInP,
\end{align*}
where $\Delta T$ is the uniform timestep between trajectory states.

\subsection{Walking model}

\begin{figure}[tb!]
\begin{centering}
\includegraphics[width=0.3\textwidth]{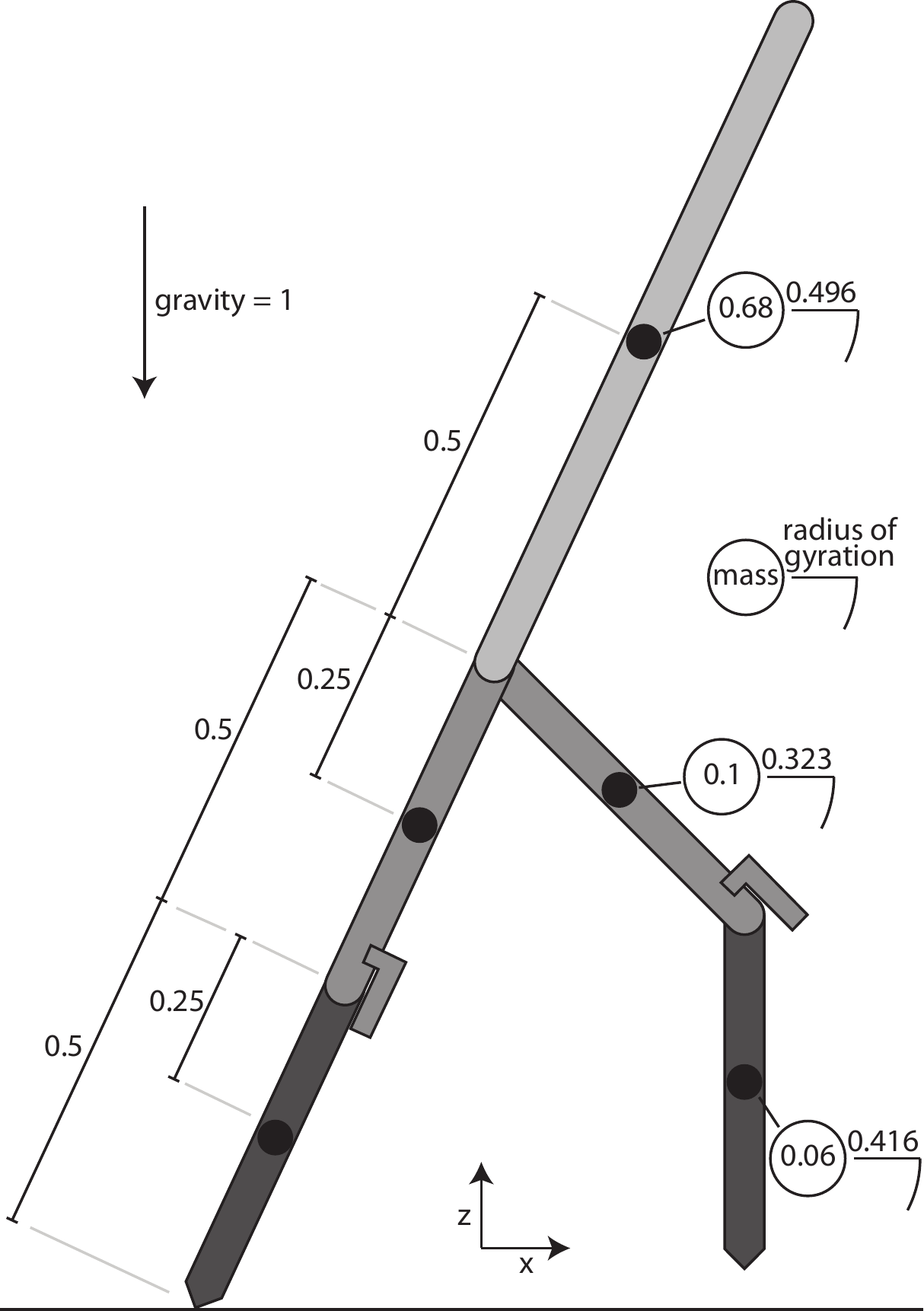}
\vspace{2em}
\caption{The model used in this paper. It is based on a ballistic walker \cite{mochon_ballistic_1980} with added torso. The knee stops can provide a stopping force to avoid knee hyperextension, though the knee does not lock. The feet are perfectly inelastic point contacts. Mass distributions are roughly anthropomorphic. There are 7 degrees of freedom, three for the floating torso, two hips, and two knees. All units are dimensionless using the non-dimensionalizing quantities leg length, gravity, and body mass.} \label{fig:modelDescription}
\end{centering}
\vspace{-1.5em}
\end{figure}

We consider a simple walking model based on a ballistic walker \cite{mochon_ballistic_1980}. Our implementation consists of two thigh and two shank segments, connected by knees that keep the leg from hyper-extending. We augment the ballistic walker with a torso \refFig{fig:modelDescription} to model a lumped mass upper body. This model omits many aspects of human locomotion, such as muscles, ligaments, and detailed anatomical joints. However, we seek to capture the \edit{role} of the major joints involved in walking, which are often analyzed in terms of overall rotational motion and simple torques.
\edit{However, while this model contains many aspects of gait, such as a heavy swinging leg, ground impacts, and torso balancing, we consider it simply representative of walking models. Models should be chosen based on the experimental protocol and hypothesis under consideration.}

We use a five segment planar walker \figRef{fig:modelDescription}. Each leg is comprised of thigh and shank segments, connected at a knee pin joint. The top of each thigh is connected with a hip pin joint to a torso segment. The mass and inertia of each segment are representative of human properties \cite{winter_biomechanics_2009}.
The knee and hip joints are powered by torque actuators. In addition, a pin joint is added between the ground and the stance foot. Constraints ensure the ground cannot pull downward on the foot. The stance foot is constrained to leave the ground at the same time the swing foot lands. To ensure the ground reaction force does not accelerate the foot off the ground, the foot velocity when leaving the ground is zero. Additionally, a constraint disallows knee hyperextension. Simply adding this constraint may require negative work from the knee actuator to avoid hyperextension. However, a human has a system of bone and ligaments which passively apply force to resist hyperextension. To simulate this, a kneelock torque is added to each leg, allowing a unilateral holding torque with no associated actuation cost. Kneelock torque is only allowed at full extension, and cannot add energy to the system.
The cost components considered are
\begin{align*}
&\phi_{\textrm{kneeTorque}} = \sum_t (\tau_{\textrm{knee, swing, t}}^2 + \tau_{\textrm{knee, stance, t}}^2) \Delta T \hspace{1em} \textrm{and}\\
&\phi_{\textrm{hipTorque}} = \sum_t (\tau_{\textrm{hip, swing, t}}^2 + \tau_{\textrm{hip, stance, t}}^2) \Delta T, \forall \tInT \\
&\phi_{\dot{\textrm{torque}}} = 0.001 * \sum_t
\biggl(
\left(\frac{\tau_{\textrm{hip, swing, t+1}} - \tau_{\textrm{hip, swing, t}}}{\Delta T} \right)^2 \Delta T + \\
&\left(\frac{\tau_{\textrm{hip, stance, t+1}} - \tau_{\textrm{hip, stance, t}}}{\Delta T} \right)^2 \Delta T + \\
&\left(\frac{\tau_{\textrm{knee, swing, t+1}} - \tau_{\textrm{knee, swing, t}}}{\Delta T} \right)^2 \Delta T + \\
&\left(\frac{\tau_{\textrm{knee, stance, t+1}} - \tau_{\textrm{knee, stance, t}}}{\Delta T} \right)^2  \Delta T,
\forall \tInTMinusOne
\end{align*}
We scale $\phi_{\dot{\textrm{torque}}}$ by 0.001 due to unit differences. With more cost components with different scales, a more sophisticated regularization scheme would likely help the numerical properties of the algorithm.
\edit{These cost components are not exhaustive, rather they represent generic effort costs that may be used both in biomechanics and robotics contexts.}

\subsection{Trajectory Optimization}

The constraint and cost functions are generated symbolically, which are in turn used to calculate sparse derivatives to attain expressions for their Jacobians and Hessians. The optimization is solved using direct transcription and an implementation of sequential quadratic programming, IPOPT \cite{wachter_implementation_2005}. The initial conditions are a rough approximation of a kinematically repetitive gait, but without regard to dynamic feasibility or specific constraint satisfaction. Small amounts of noise are added to the initial condition and the algorithm is run 15 times in parallel with the best feasible solution found chosen. This was found to better avoid inferior local minima. \edit{While in principle other constrained nonlinear program solver could be used, we have found IPOPT sufficiently robust and performant for the present study.}
\edit{Similar models can be optimized using standard MATLAB code \cite{kelly_introduction_2017}.}

\begin{figure*}[bth!]
\begin{center}
\centering
\includegraphics[width=0.95\textwidth]{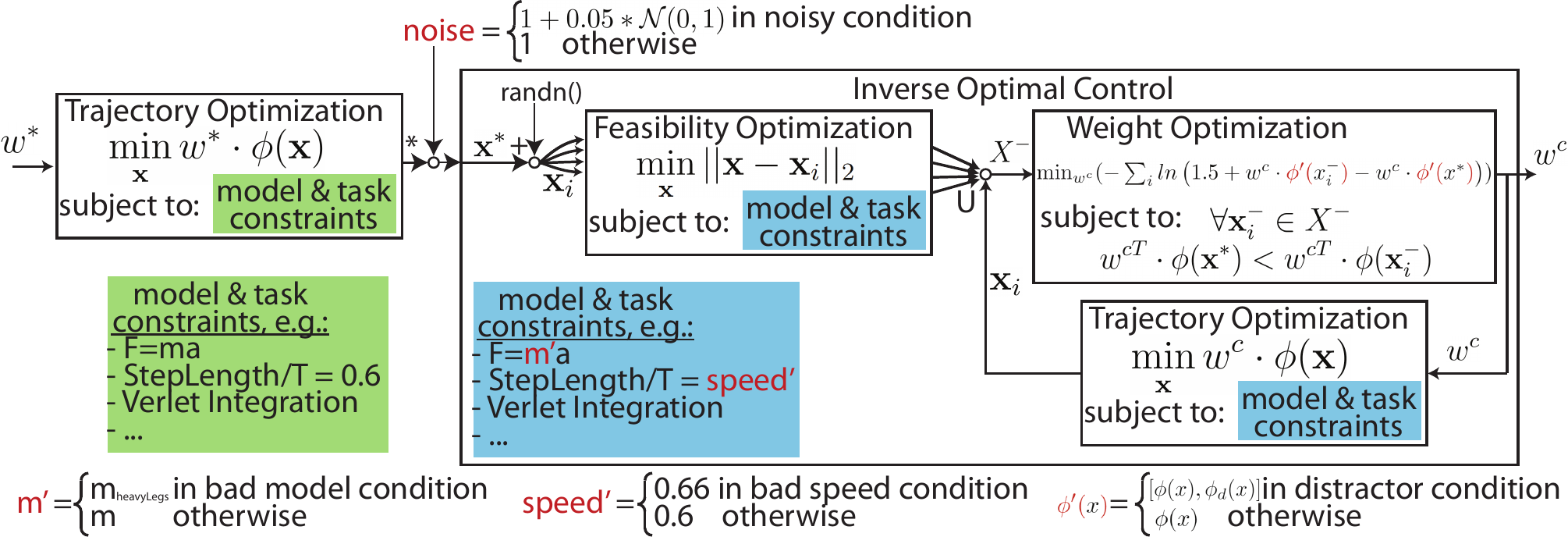}
\vspace{-0.5em}
\caption{
The \IOC procedure and test conditions. A trajectory optimization (left) generates a trajectory using ground truth cost weights $w^*$. Trajectories are then randomly generated around this observed trajectory, and then constrained to be feasible. These feasible trajectories are treated as suboptimal, and a weight optimization is then performed to produce a candidate set of weights which maintain the optimality of the observed optimal trajectory $x^*$ with respect to the suboptimal set $X^{-1}$. A new trajectory optimization is then performed using these candidate weights to yield a new candidate trajectory $x_i$, which we then add to the suboptimal states. We then iterate until either 25 iterations are reached, the weight candidates converge, or an optimization fails.
}
\label{fig:methodology}
\vspace{-1.5em}
\end{center}
\end{figure*}

\subsection{Idealized \IOC of simulated walking}

\IOC attempts to recover a cost function consistent with the optimality of some observed behavior. Here, we perform an idealized test of our procedure using the walking model in ideal conditions, namely that the IOC procedure has access to both the exact model and cost components used to generate the ideal \edit{trajectory} \figRef{fig:methodology}.
We \edit{represent} the observed behavior using a single demonstrated step ($x^*$), and the cost function as a weighted sum of the cost components. \edit{In addition, we test the IOC under ideal conditions with a faster gait (1.0m/s).}

We proceed by generating $N$ normally distributed trajectories around the optimal trajectory (\trajOpt), yielding a set of suboptimal, likely infeasible trajectories ($x_i^r, i \in [1 \ldots N]$) \figRef{fig:methodology}. We then project these trajectories into the feasible space by performing a trajectory optimization with the full model and task constraints to find a feasible trajectory ($x_i^{r,f}$) as close as possible to the randomly generated one by minimizing the error $|| x_i^{r,f} - x_i^r ||_2$ \figRef{fig:methodology}.
This produces a set ($X^-$) of likely suboptimal feasible trajectories.
We then find a set of weights (\weightCandidate) which is consistent with the proposition that the optimal example trajectory has a lower cost than each of the suboptimal trajectories in $X^-$. We optimize
\begin{align*} %
&\min_{\weights^{c}} -\sum_{\xVect^- \in X^-} -log(1.5 + \weights^{c} \cdot \phi(\xVect^-) - \weights^{c} \cdot \phi(\trajOpt)) \\
&\textrm{such that} \hspace{1em} \weights^{c} \cdot \phi(\trajOpt) < \weights^{c} \cdot \phi(\xVect^-), \forall \xVect^- \in X^- \\
& \hspace{4.5em} ||\weights^c||_2 = 1, \hspace{0.5em} \weights^{c} < \vec{0.999}, \hspace{0.5em} \weights^{c} > \vec{0.001}
\end{align*}
The first set of constraints require that, assuming the weight vector $\weights^{c}$, each suboptimal trajectory has a higher cost than the example trajectory. In addition, the weight vector is constrained to have unity 2-norm, as well as component-wise limits between 0.001 and 0.999 to help avoid degenerate results. The cost function prefers weights which differentiate as much as possible the cost of the optimal from the suboptimal states, nonlinearly prioritizing states with a cost closest to the optimal \figRef{fig:methodology}.
\edit{This nonlinear prioritization of trajectories close to optimal was empirically found to provide better convergence than a simple summation of cost differences.}

The candidate weight vector is then used in a trajectory optimization using the model and task constraints to generate a new stepping trajectory ($x_i^{-}$) which we assume to be suboptimal \figRef{fig:methodology}. This suboptimal trajectory is then added to $X^{-}$, and another set of weights can be calculated. This process is iteratively repeated 25 times or until the optimization fails to find a solution, typically because the weight optimization is no longer able to find a set of weights which differentiates all of the suboptimal trajectories from the optimal. We found that 25 iterations was sufficient to achieve convergence in our testing. Finally, the IOC algorithm output is the set of weights found during the iterative process which resulted in the $\xVect^-$ with the minimal norm error of positions, velocities, and joint torques from the observed optimal trajectory.

The logarithmic nonlinearity in the cost function of the weight optimization devalues suboptimal costs which are far from the optimum. This ensures that as trajectories cluster around the optimum in later iterations, their error contributions are not overpowered by the poor trajectories found in earlier iterations.
The cost chosen here improve convergence properties, but our results are not very sensitive to this cost.

\subsection{Robustness to imperfect modelling and measurements}

The method described above for testing the \IOC procedure is an important first step to determine the ability of the IOC procedure for determining cost functions. However, this test occurs under ideal conditions, when the optimal observed \edit{walking} behavior can be perfectly reproduced within the IOC.

To test the robustness of the methodology proposed to imperfections, we performed IOC on the walking model with
four types of error \figRef{fig:methodology}. The first is additive Gaussian white noise ($\mu$: 0, $\sigma$: 5\% of signal amplitude) on the measured trajectory. This tests the robustness of the IOC results to pure measurement noise. Then, to test the effect of model error, we assume a model with 10\% more mass in each leg segment than that used to generate the optimal data. Mass is removed from the torso to maintain a total mass of one.
To test imperfect knowledge of the task, we test \IOC using a model constrained to walk 10\% faster than the observed trajectory.
To consider imperfect cost component knowledge, we add three distractors costs to the set of cost components within the IOC: the integrated 2-norm of the torso angle, step length error from a desired value of 0.3, and the mean swing foot height. %

\section{Results}

\edit{The trajectory optimization reliably finds local optima for each weight condition tested. Furthermore, in each feature penalizing case, the penalized feature is lower than in the other conditions. This shows the optimization finds behaviors which tradeoff the various costs appropriately (Table \ref{table:features})}
\begin{table}[]
\begin{tabular}{ccccc}
& nominal   & penalize: $\tau_{\mathbf{hip}}$ &  $\tau_{\mathbf{knee}}$ &  $\dot{\tau}$ \\
\cline{2-5}
$\phi_{\tau_{\mathbf{hip}}}$               & 0.0975 & 0.0360                             & 0.192                               & 0.112                                            \\
$\phi_{\tau_{\mathbf{knee}}}$              & 0.0809 & 0.204                              & 0.0331                              & 0.0897                                           \\
$\phi_{\dot{\tau}}$ & 0.0142 & 0.0351                             & 0.0206                              & 0.00307                                          \\
\end{tabular} \caption{Features found for the optimal trajectories across the different penalizing conditions.}
\vspace{-2em}
\label{table:features}
\end{table}

\begin{figure}[thpb]
\centering
\includegraphics[width=0.425\textwidth]{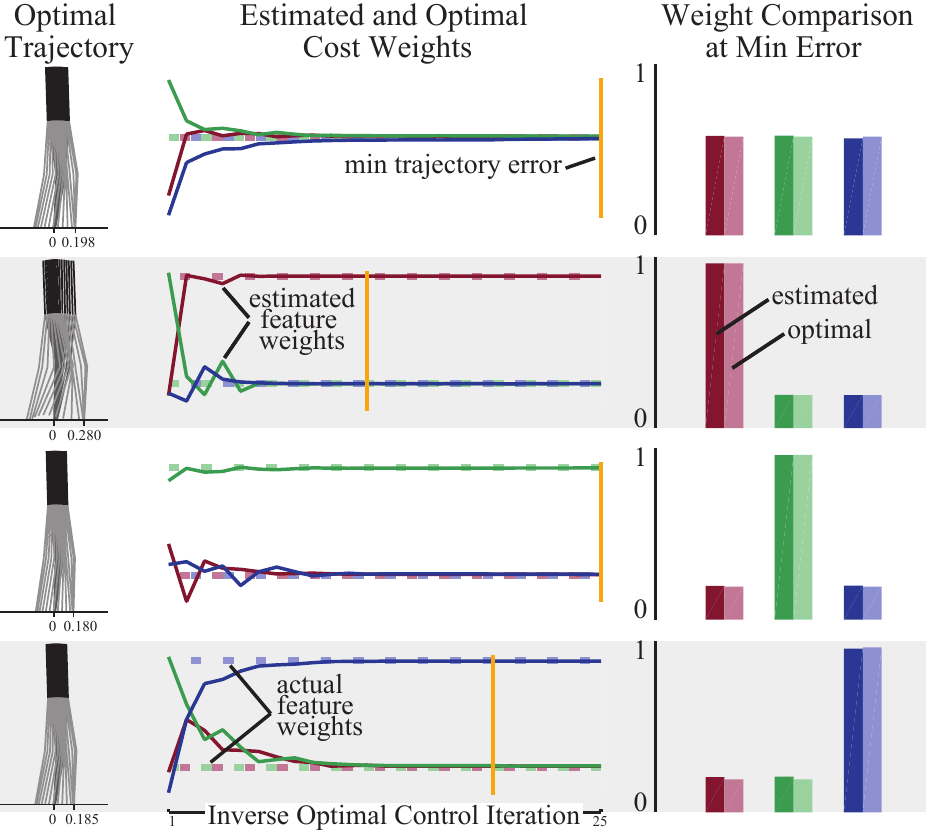}
\caption{Results of the \IOC algorithm under optimal conditions. The optimal trajectory found is in the left column. To the right are shown the estimated optimal cost weights throughout the process of the \IOC. The dashed horizontal lines show the correct cost weights, the solid lines show the candidate weights ($w^c$) throughout the iterative process. Also included is a light gold line representing the error of the trajectory generated using the candidate weights of that iteration. Lastly are shown the estimated weights, which are the candidate weights measured at the iteration with minimal trajectory error throughout the \IOC process.
} \label{fig:iocResults}
\vspace{-1em}
\end{figure}

\begin{figure*}[thpb]
\centering
\includegraphics[width=0.9\textwidth]{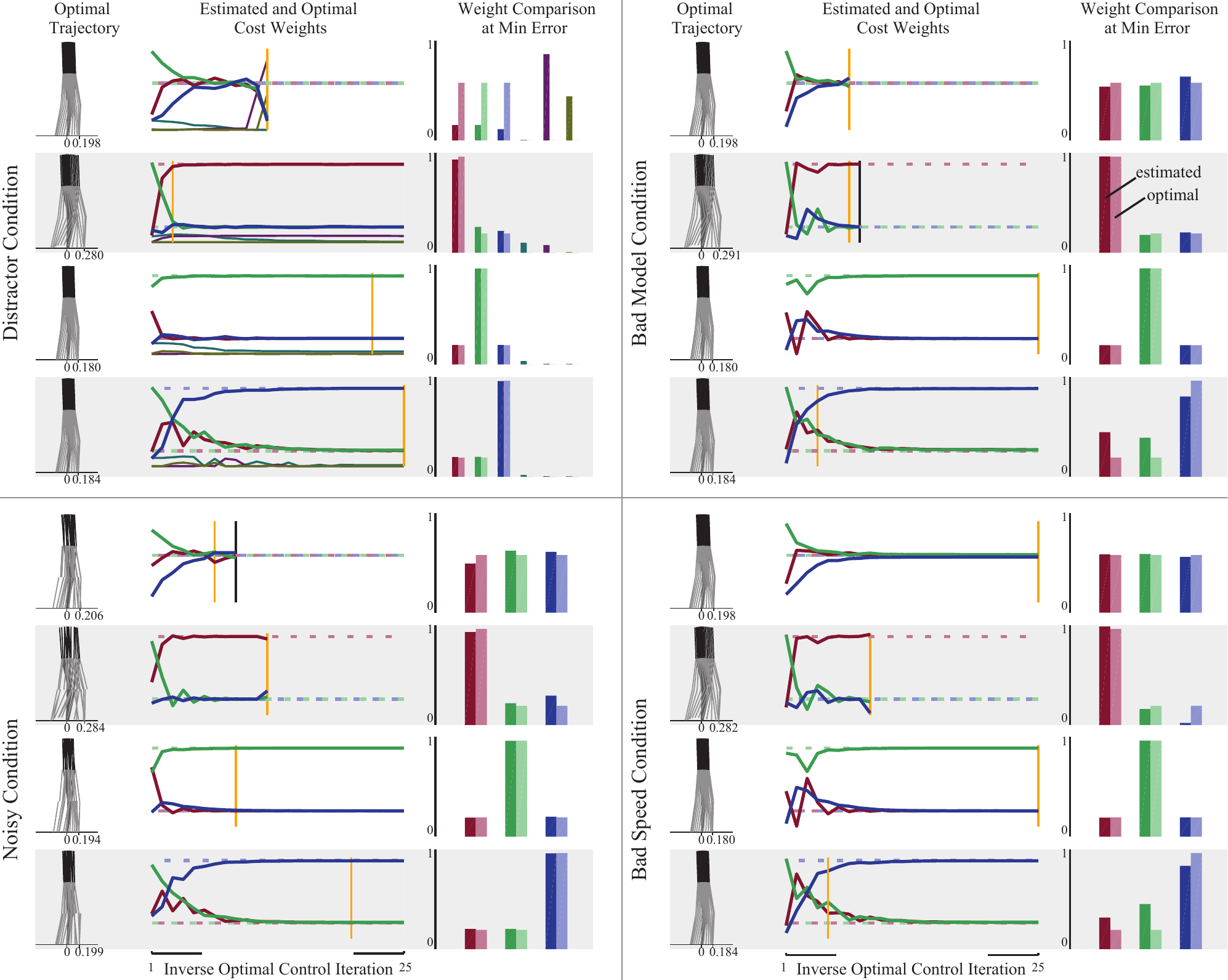}\caption{
\edit{IOC results under a variety of imperfect conditions. (top left) The distractor condition adds 3 additional cost components inside the \IOC which were not used in the original optimization. For the nominal condition (1st row) the IOC finds optimal gains with a substantial gain for the distractor costs associated with torso angle and foot height. However, note the preceding iterations with weight close to optimal.
(top right) The bad model condition uses a walker model with 10\% more mass in the leg segments than the model that was used to generate the optimal trajectory. (bottom left) The noisy condition has simple additive Gaussian noise with a standard deviation of 5\% of the span of the signal added to each signal in the observed trajectory. Note the noise apparent in the angles of the optimal trajectory. (bottom right) The bad speed condition uses a speed constraint of 10\% higher in the model used in the IOC than that used to generate the optimal walking behavior.}
} \label{fig:iocResultsImperfectCases}
\end{figure*}

Under ideal conditions, the IOC algorithm identifies the strongest component for each set of weights tested \figRef{fig:iocResults}. With all components equal, the three weights are approximately equal.
When including distractor factors \figRefAnnotate{fig:iocResultsImperfectCases}{, top left},
\edit{only the nominal weighting results in a substantial amount of weight attributed to distractor features.}
The other distractor cases are qualitatively similar to the ideal condition.

\edit{The added noise condition was qualitatively similar to the ideal case \figRefAnnotate{fig:iocResultsImperfectCases}{, bottom left}. The incorrect speed \figRefAnnotate{fig:iocResultsImperfectCases}{, bottom right} and incorrect mass \figRefAnnotate{fig:iocResultsImperfectCases}{, top right} conditions also both correctly identify prioritized components. Note the similarities in convergence properties between these two conditions. Each condition converges toward the correct weights, with the worst cases due to a pre-convergence trajectory having lower error than the correct weights.}

\begin{figure}[thpb]
\centering
\includegraphics[width=0.48\textwidth]{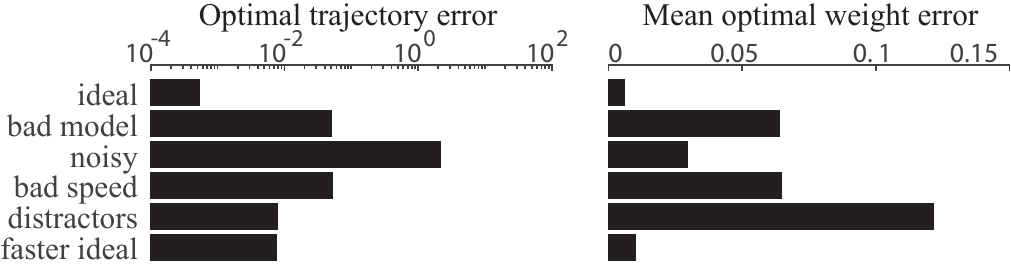}
\vspace{-1em}
\caption{Total errors of the IOC estimate across conditions. (left) We measure weight estimation error as the average absolute difference between the estimated weights and the optimal weights. \edit{(right)
Trajectory error is the sum of the squared difference of each segment angle and angular velocity of the best found trajectory from the observed optimal.}
} \label{fig:summary}
\vspace{-1em}
\end{figure}

To summarize the performance of the IOC algorithm we calculate the average of the absolute difference between the estimated and optimal weights, \figRefAnnotate{fig:summary}{, left}.
\edit{
The ideal conditions both have less than 1\% mean weight estimation error while the distractor condition has the worst effect on weight estimation. We also consider the optimal trajectory error \figRefAnnotate{fig:summary}{, right}, which shows the ability of the IOC to regenerate the kinematics of the optimal trajectory. Here the noisy condition has the highest error, and the ideal, distractor, and faster ideal conditions have an order of magnitude less error than the other perturbed cases.
}

\section{Discussion}

We find that under ideal circumstances the \IOC procedure finds the optimal gains \figRef{fig:iocResults}. Generally the results converge to the final results within 10s of iterations. We consider this to be an acceptable convergence rate, albeit with only a few cost components considered. \edit{IOC found weights and trajectories close to optimal in both the nominal gait and the fast gait under ideal conditions \figRef{fig:summary}.}

Results vary more with the imperfect IOC tests.
\edit{IOC has difficulty reliably recovering cost weights in the presence of distractor costs. The iterative process considers weights close to optimal, however a slightly lower trajectory error is found with substantial weight assigned to distractor costs \figRefAnnotate{fig:iocResultsImperfectCases}{, top left}. We see that a linear combination of distractor costs (step length difference and foot height) is difficult to disambiguate from the actual costs in this condition. However the optimal trajectory error is on the same order as the faster ideal condition, suggesting that the weights found are roughly as valid as the optimal weights. An L1 norm error term on the weights could be used to find sparse cost functions consisting of as few terms as possible while still explaining the data.}

This illustrates a general issue with IOC, which is the non-uniqueness of solutions. For example, we might choose joint torque squared as a cost component, but also torque cubed. As these components are highly correlated, IOC may struggle to reliably disambiguate between them given a single trial. A more subtle redundancy arises when using a cost for force fluctuation based on metabolic costs in muscles, in addition to a conventional jerk minimizing cost \cite{hogan_organizing_1984}. Since these are roughly linearly related by the dynamics equations ($\mathbf{force}$ $\alpha$ $\mathbf{acceleration}$ $\rightarrow$ $\dot{\mathbf{force}}$ $\alpha$ $\mathbf{jerk}$), disambiguating between torque derivative and jerk is challenging. However, there may be experiments
which help disambiguate these costs.

We found the IOC algorithm was more robust to other artificially introduced imperfections. Simply adding Gaussian noise to the measured signal had little effect  \figRefAnnotate{fig:iocResultsImperfectCases}{, lower left}.
\edit{The trajectory error is the highest of any tested condition, but the weight error is lowest of any perturbed condition. Although the IOC cannot reproduce the noisy trajectory, it is able to infer the underlying cost function.}
This may be because the noisy signal is generally infeasible, but the trajectory is fixed early in the IOC process. Recall that the first step of the IOC finds feasible noisy versions of the optimal trajectory.
Enforcing feasibility likely smooths the noisy signal, which may impart some noise robustness. It is possible that the observed robustness to the mass and treadmill speed disturbances is the result of a similar effect, as changing these aspects of the model will result again in an infeasible observed input trajectory. Overall the results demonstrate qualitative robustness to the types of imperfections tested. However, as the weight optimization compares smoothed feasible trajectories to the original measured trajectory, imperfections in the original trajectory may still be mistaken for preference if they can be explained by a cost component. One possible solution to a particularly poor input signal is to find the closest feasible trajectory using the trajectory optimization before performing IOC on it.
\edit{These robustness properties suggest application to human and robot experiments may be appropriate.}

\subsection{Limitations and possible solutions}

One significant limitation of this method is the strong dependence on model choice. For example, some costs humans consider may be based on the properties of particular muscles, connective tissue, and joints. The model discussed in this paper does not include detailed anatomical models of natural muscles, and therefore would have difficulty capturing such detailed costs. However, as robots likely have different underlying actuation details \edit{from humans}, a detailed cost function based on human muscles would have limited applicability to robots. A more abstract cost function may be more portable to machines which are human-like but with different underlying actuation. In addition, there may be fundamental considerations such as stability that are independent of the actuation details.
\edit{While trajectory optimization of more complex models might be more difficult, it is important to note that there exists very efficient optimizers capable of handling more complex humanoid dynamics \cite{herzog_structured_2016, clever_inverse_2016} and muscle modelling \cite{wang_optimizing_2012}.}

There is also a more subtle difficulty in distinguishing preference from requirements. As an optimal controller considers both constraints and cost, if a chosen model fails to account for constraints in the observed task, those constraints may be treated instead as preference in the IOC. On the other hand, if a constraint is added to the model which is not present in the observed task, the model may either have difficulty reproducing the original trajectory, or would fail to account for the incorrectly constrained behavior in the cost weights. A potential solution is a more comprehensive approach to constraints, where a family of models is tested with different combinations of constraints, with a range of experiments to test for consistency of the inferred cost weights across conditions.

Finally, the cost components chosen are also a matter of model selection with a strong effect on the outcome of the algorithm. As noted previously, this process can only discover combinations of the given cost components consistent with the observed behavior, rather than generate the correct cost function. It is likely when analyzing experimental human data that unmodelled cost components exist which better represent a subject's preference. As demonstrated with the distractor cost test, care must be taken to choose cost components that have separable effects on the optimal trajectory. As many sets of cost components may allow a model to reproduce an observed behavior, choosing the components may be considered a hypothesis of possible underlying costs. Potential approaches to hypothesis development might include leave-one-out testing of cost components, with performance determined by consistency of cost functions across the different conditions under study. Leaving out some components may reduce the consistency of the cost weights found across conditions more than others. Those with a strongest effect on consistency might be considered to be the most useful (in the context of the rest of the components) for describing the underlying cost function.

\subsection{Further Work}
While our approach will infer cost functions from existing experimental human data, IOC is also a promising tool for analyzing subject preference in new experimental designs as well. For example, comparing the preference tradeoffs between stability and energetic efficiency in different patient populations may help provide insight into different medical conditions. Furthermore, a cost function found using IOC may serve as an identifying signature of individual differences between individuals affected by unique injuries or conditions.

\edit{The robustness analysis procedure presented can be applied to other models. We plan to extend this model with ankles to allow calculation of standard metrics of planar hip, knee, and ankle kinematics and torques \cite{zelik_human_2010}. Additional extensions such as to three dimensions could allow treatment of additional concepts such as lateral stability.}
\edit{Since the trajectory optimization on which IOC depends is inherently a local search, it is difficult to extrapolate a verification of IOC procedure to very different models or tasks. Our results suggest that IOC properly identifies weight changes around the tested behavior, but each model should be verified per behavior.}

The results of IOC studies can also be used to generate new behavior. A generative model with a cost function can synthesize new behavior given new task constraints to test generalizability.

For example, \IOC on walking data from a human subject yields a cost function which could then generate modelled stair walking behavior. The stair walking model gait could then be compared to data of the same subject walking up stairs. %
\edit{As we do not propose to find a universal cost function for all human behavior, such generalization tests would illustrate the scope of a given cost function.}
\edit{As with standard model fitting techniques, IOC can be applied to different experimental questions by considering more than a single trial. IOC can estimate a cost function for multiple trials, limiting the effects of noise and variability. Also, considering multiple conditions can improve the scope of the resulting cost function, or considering multiple subjects can yield an average cost function for comparison with other populations.}

\section*{Acknowledgments}
{\footnotesize{
Any opinions, findings, and conclusions or recommendations expressed in this material are those of the authors and do not necessarily reflect the views of the funding organizations.
}}

\bibliographystyle{IEEEtran}
\bibliography{IEEEabrv,jrebulaLibrary}

\end{document}